\documentclass[11pt,a4paper]{article}
\usepackage[hyperref]{naaclhlt2019}
\usepackage{times}
\usepackage{latexsym}
\usepackage{hyperref}
\usepackage{url}

\usepackage{graphicx, subfigure}
\graphicspath{ {./images/} }
\usepackage{stfloats}
\usepackage{float}

\usepackage{tabularx, booktabs}
\newcolumntype{Y}{>{\centering\arraybackslash}X}

\usepackage{url}

\aclfinalcopy 
\def\aclpaperid{***} 
\usepackage{mathtools}
\usepackage{amsmath}
\usepackage{amssymb}
\usepackage{bbm}
\usepackage{bm}

\DeclareMathOperator*{\metric}{metric}
\DeclareMathOperator*{\intra}{intra}
\DeclareMathOperator*{\inter}{inter}

\DeclarePairedDelimiter\abs{\lvert}{\rvert}%
\DeclarePairedDelimiter\norm{\lVert}{\rVert}%

\usepackage{amsthm}

\usepackage{multirow}

\title{Mitigating Uncertainty in Document Classification}

\author{Xuchao Zhang{$^\dag{}^\ddag$}, Fanglan Chen{$^\dag$}, Chang-Tien Lu{$^\dag$}, Naren Ramakrishnan{$^\dag$} \\ 
	{$^\dag$}Discovery Analytics Center, Virginia Tech, Falls Church, VA, USA\\
	{$^\ddag$}NEC Laboratories America, Inc, Princeton, NJ, USA\\
	{$^\dag$}\{xuczhang, fanglanc, ctlu, naren\}@vt.edu, {$^\ddag$}xuczhang@nec-labs.com}
\date{}

\begin{document}
\maketitle
\begin{abstract}
The uncertainty measurement of classifiers' predictions is especially important in applications such as medical diagnoses that need to ensure limited human resources can focus on the most uncertain predictions returned by machine learning models. However, few existing uncertainty models attempt to improve overall prediction accuracy where human resources are involved in the text classification task. In this paper, we propose a novel neural-network-based model that applies a new dropout-entropy method for uncertainty measurement. We also design a metric learning method on feature representations, which can boost the performance of dropout-based uncertainty methods with smaller prediction variance in accurate prediction trials. 
Extensive experiments on real-world data sets demonstrate that our method can achieve a considerable improvement in overall prediction accuracy compared to existing approaches. In particular, our model improved the accuracy from 0.78 to 0.92 when 30\% of the most uncertain predictions were handed over to human experts in ``20NewsGroup" data.
\end{abstract}

\section{Introduction}

Machine learning algorithms are gradually taking over from the human operators in tasks such as machine translation \cite{bahdanau2014neural}, optical character recognition \cite{mithe2013optical}, and face recognition \cite{parkhi2015deep}. However, some real-world applications require higher accuracy than the results achieved by state-of-the-art algorithms, which makes it difficult to directly apply these algorithms in certain scenarios. For example, a medical diagnosis system \cite{van2017bayesian} is expected to have a very high accuracy to support correct decision-making for medical practitioners. Although domain experts can achieve a high performance in these challenging tasks, it is not always feasible to rely on limited and expensive human input for large-scale data sets. Therefore, if we have a model with 70\% prediction accuracy, it is intuitive to ask what percentage of the data should be handed to domain experts to achieve an overall accuracy rate above 90\%? 
To maximize the value of limited human resources while achieving desirable results, modeling uncertainty accurately is extremely important to ensure that domain experts can focus on the most uncertain results returned by machine learning models. 
%


Most existing uncertainty models are based on Bayesian models, which are not only time-consuming but also unable to handle large-scale data sets. 
Deep Neural networks (DNNs) have attracted increasing attention in recent years and have been reported to achieve state-of-the-art performance in various machine learning tasks \cite{yang2016hierarchical,iyyer2014neural}. However, unlike probabilistic models, DNNs are still at the early development stage in regards to providing the model uncertainty in their predictions. For those seeking to address the prediction uncertainty in DNNs, it is common to suffer from the following issues on the text classification task. 
Firstly, few researchers have sought to improve overall prediction performance when only limited human resources are available. Different from existing methods which focus on the value of uncertainty, this problem needs to get domain experts involved in emphasis on the order of the uncertain predictions. 
For example, the importance of distance between feature representations is neglected by the majority of existing models, but actually this is crucial for improving the order of uncertain predictions, especially during the pre-training of embedding vectors.
Moreover, the methods proposed for continuous feature space cannot be applied to discrete text data. For example, adversarial training is used in some uncertainty models \cite{goodfellow2014explaining, lakshminarayanan2017simple, mandelbaum2017distance}. However, due to its dependence on gradient-based methods to generate adversarial examples, the method is not applicable to discrete text data.


In order to simultaneously address all these problems in existing methods, the work presented in this paper adopts a DNN-based approach that incorporates a novel dropout-entropy uncertainty measurement method along with metric learning in the feature representation to handle the uncertainty problem in the document classification task.
The study's main contributions can be summarized as follows:
\begin{itemize}
\item A novel DNN-based text classification model is proposed to achieve higher model accuracy with limited human input. In this new approach, a reliable uncertainty model learns to identify the accurate predictions with smaller estimated uncertainty.   
\item Metric learning in feature representation is designed to boost the performance of the dropout-based uncertainty methods in the text classification task. Specifically, the shortened intra-class distance and enlarged inter-class distance can reduce the prediction variance and increase the confidence for the accurate predictions. 
\item A new dropout-entropy method based on the Bayesian approximation property of Dropout in DNNs is presented. Specifically, we measure the model uncertainty in terms of the information entropy of multiple dropout-based evaluations combined with the de-noising mask operations.
\item Extensive experiments on real-world data sets demonstrate that the effectiveness of our proposed approach consistently outperforms existing methods. In particular, the macro-F1 score can be increased from 0.78 to 0.92 by assigning 25\% of the labeling work to human experts in a 20-class text classification task.
\end{itemize}

The rest of this paper is organized as follows. Section \ref{section:related_work} reviews related work, and Section \ref{section:model} provides a detailed description of our proposed model. The experiments on multiple real-world data sets are presented in Section \ref{section:experiment}. The paper concludes with a summary of the research in Section \ref{section:conclusion}.

\section{Related Work} \label{section:related_work}
The work related to this paper falls into two sub topics, described as follows.
\subsection{Model Uncertainty}
Existing uncertainty models are usually based on Bayesian models, which is Traditional Bayesian models such as Gaussian Process (GP), can measure uncertainty of model. However, as a non-parametric model, the time complexity of GP is increased by the size of data, which makes it intractable in many real world applications. 

Conformal Prediction (CP) was proposed as a new approach to obtain confidence values \cite{vovk1999machine}. Unlike the traditional underlying algorithm, conformal predictors provide each of the predictions with a measure of confidence. Also, a measure of ``credibility” serves as an indicator of how suitable the training data are used for the classification task \cite{shafer2008tutorial}. Different from Bayesian-based methods, CP approaches obtain probabilistically valid results, which are merely based on the independent and identically distributed assumption. The drawback of CP methods is their computational inefficiency, which renders the application CP not applicable for any model that requires long training time such as Deep Neural Networks.

With the recently heated research on DNNs, the associated uncertainty models have received a great deal of attention. 
Bayesian Neural Networks are a class of neural networks which are capable of modeling uncertainty \cite{Denker:1990:TNO:2986766.2986882} \cite{hernandez2015probabilistic}. These models not only generate predictions but also provide the corresponding variance (uncertainty) of predictions. However, as the number of model parameters increases, these models become computationally more expensive \cite{wang2016towards}. 
Lee et al. proposed a computationally efficient uncertainty method that treats Deep Neural Networks as Gaussian Processes \cite{lee2017deep}. Due to its kernel-based design, however, it is not straightforward to apply this to the deep network structures for text classification.
Gal and Ghahramani used dropout in DNNs as an approximate Bayesian inference in deep Gaussian processes \cite{Gal:2016:DBA:3045390.3045502} to mitigate the problem of representing uncertainty in deep learning without sacrificing the computational complexity. Dropout-based methods have also been extended to various tasks such as computer vision \cite{kendall2017uncertainties}, autonomous vehicle safety \cite{mcallister2017concrete} and medical decision making \cite{van2017bayesian}. 

However, few of these methods are specifically designed for text classification and lack of considerations on improving the overall accuracy in the scenario that domain experts can be involved in the process. 


\subsection{Metric Learning}
Metric learning \cite{xing2003distance, weinberger2006distance} algorithms design distance metrics that capture the relationships among data representations. This approach has been widely used in various machine learning applications, including image segmentation \cite{gong2013fuzzy}, face recognition \cite{guillaumin2009you}, document retrieval \cite{Xu:2012:SDM:2396761.2398536}, and collaborative filtering \cite{hsieh2017collaborative}. 
Weinberger et al. proposed a large margin nearest neighbor (LMNN) method \cite{weinberger2006distance} 
in learning a metric to minimize the number of class impostors based on pull and push losses.
However, as yet there have been no report of work focusing specifically on mitigating prediction uncertainties. Mandelbaum and Weinshall \cite{mandelbaum2017distance} measured model uncertainty by the distance when comparing to the feature representations in training data, but this makes the uncertainty measurement inefficient because it requires an iteration over the entire training data set.
To the best of our knowledge, we are the first to apply metric learning to mitigate model uncertainty in the text classification task. We also demonstrate that metric learning can be applied to dropout-based approaches to improve their prediction uncertainty.

\section{Model} \label{section:model}

In this section, we propose a DNN-based approach to predict document categories with high confidence for the accurate predictions and high uncertainty for the inaccurate predictions. The overall architecture of the proposed model is presented in Section \ref{section:model_overall}. The technical details for the metric loss and model uncertainty predictions are described in Sections \ref{section:metric_loss} and \ref{section:model_uncertainty}, respectively.

 

\begin{figure}
\centering
  \includegraphics[scale=0.27]{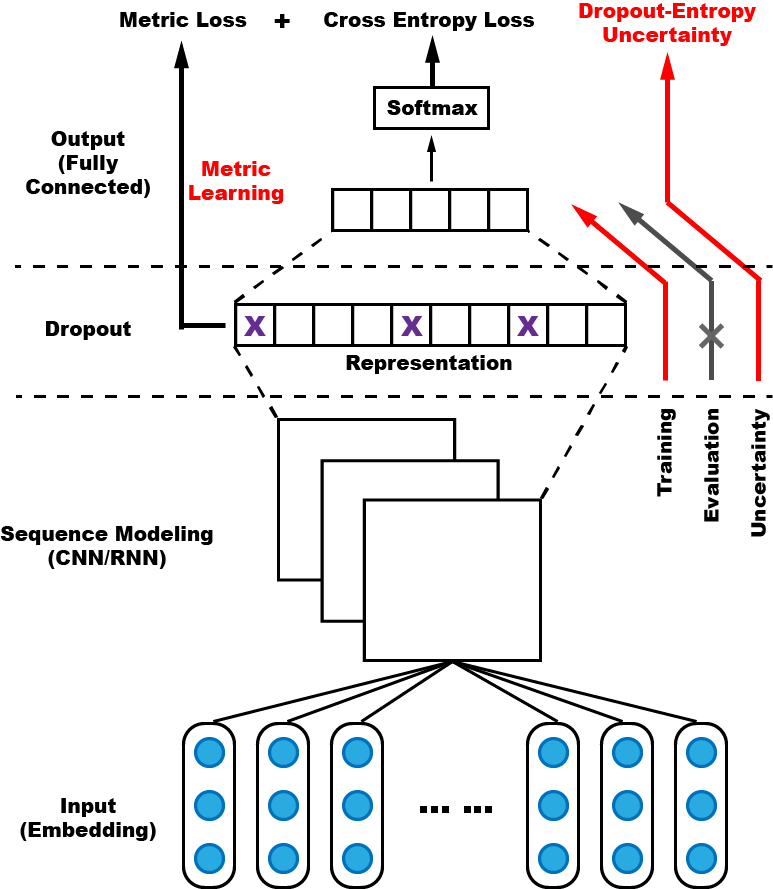}
  \caption{\small Overall Architecture of Proposed Model}
  \label{fig:model}
\end{figure}



\begin{figure*}
\centering
  \includegraphics[scale=0.425]{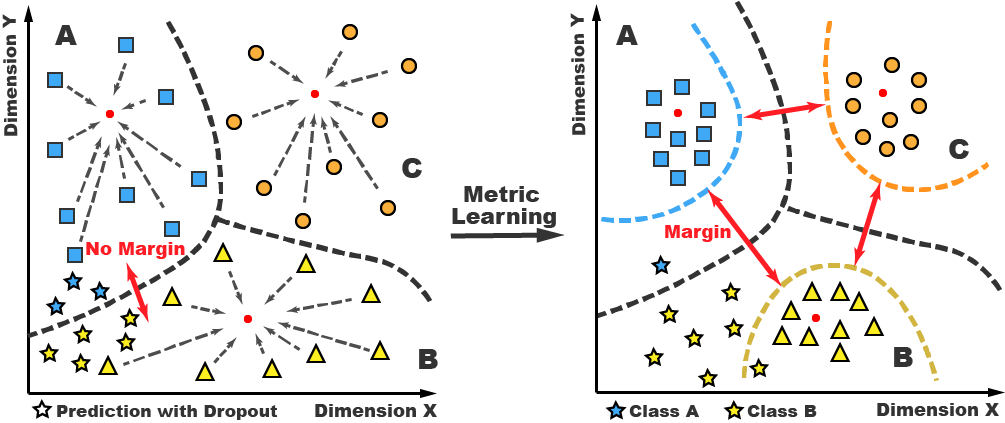}
  \caption{\small Feature representations with no metric learning (left) and metric learning (right).}
  \label{fig:metric}v
\end{figure*}

\subsection{Model Overview} \label{section:model_overall}
In order to measure the uncertainty of the predictions for document classification task, we propose a neural-network-based model augmented with dropout-entropy uncertainty measurement and incorporating metric learning in its feature representation. The overall structure of the proposed model is shown in Figure \ref{fig:model}. Our proposed model has four layers: 
1) \textbf{Input Layer}. The input layer is represented by the word embeddings of each words in the document. By default, all word vectors are initialized by Glove \cite{pennington2014glove} pre-trained word vectors in Wikipedia with an embedding dimension of 200. 
2) \textbf{Sequence Modeling Layer}. The sequence modeling layer extracts the feature representations from word vectors. This is usually implemented by Convolutional Neural Networks (CNN) or Recurrent Neural Networks (RNN). In this paper, we focus on a CNN implementation with max pooling that utilizes 3 kernels with filter sizes of 3, 4 and 5, respectively. After that, a max pooling operation is applied on the output of sequence model.
3) \textbf{Dropout layer}. The convolutional layers usually contain a relatively small number of parameters compared to the fully connected layers. It is therefore reasonable to assume that CNN layers suffer less from overfitting, so Dropout is not usually used after CNN layers as it achieves only a trivial performance improvement \cite{JMLR:v15:srivastava14a}. However, since there is only one fully-connected layer in our model, we opted to add one Dropout layer after the CNN layer, not only to prevent overfitting, but also to measure prediction uncertainty \cite{Gal:2016:DBA:3045390.3045502}. The Dropout operation will be randomly applied to the activations during the training and uncertainty measurement phrases, but will not be applied to the evaluation phrase. 
4) \textbf{Output layers}. The output is connected by a fully connected layer and the softmax. The loss function of our model is the combination of the cross entropy loss of the prediction and the metric loss of the feature representation. We regard the output of the Dropout layer as the representation of the document and deposit it into a metric loss function. The purpose here is to penalize large distance feature representations in the same class and small distance feature representations among different classes. The details of the metric loss function will be described in Section \ref{section:metric_loss}.

%

\subsection{Metric Learning on Text Features} \label{section:metric_loss}
For uncertainty learning in text feature space, our purpose is to ensure the Euclidean distance between intra-class instances is much smaller than the inter-class instances. To achieve this, we use metric learning to train the desirable embeddings. Specifically, let $\bm r_i$ and $\bm r_j$ be the feature representations of instances $i$ and $j$, respectively, then the Euclidean distance between them is defined as $\mathcal{D}(\bm r_i, \bm r_j) = \frac{1}{d}\norm{\bm r_i - \bm r_j}_2^2$,
where $d$ is the dimension of the feature representation.

Suppose the data instances in the training data contain $n$ classes and these are categorized into $n$ subsets $\{S_k\}_{k=1}^n$, where $S_k$ denotes the set of data instances belong to class $k$. Then the intra-class loss penalizes the large Euclidean distance between the feature representations in the same class, which can be formalized as Equation \eqref{eq:intra_loss}.
\begin{equation} \label{eq:intra_loss}
\mathcal{L}_{\intra}(k) = \frac{2}{\abs{S_k}^2 - \abs{S_k}}\sum_{i,j \in S_k, i < j} \mathcal{D}(\bm r_i, \bm r_j) \\
\end{equation}

where $\abs{S_k}$ represents the number of elements in set $S_k$. The loss is the sum of all the feature distances between each possible pair in the same class set. Then, the loss is normalized by the number of unique pairs belonging to each class set.
 
The inter-class loss ensures large feature distances between different classes, which is formally defined as Equation \eqref{eq:inter_loss}.
\begin{equation} \label{eq:inter_loss}
\mathcal{L}_{\inter}(p, q) = \frac{1}{\abs{S_p}\cdot \abs{S_q}} \sum_{i\in S_p, j \in S_q} \bigg[m - \mathcal{D}(\bm r_i, \bm r_j) \bigg]_+
\end{equation}

where $m$ is a metric margin constant to distinguish between the intra- and inter-classes and $[z]_+=\max(0, z)$ denotes the standard hinge loss. If the feature distance between instances from different classes is larger than $m$, the loss is zero. Otherwise, we use the value of $m$ minus the distance as its penalty loss, with a larger $m$ representing a larger inter-class distance. This parameter usually varies when we use different word embedding methods. In our experiment, we found that a small $m$ is normally needed when the word embedding is initialized by a pre-trained word vector method such as Glove \cite{pennington2014glove}; a larger $m$ is required if word vectors are initialized randomly.
The overall metric loss function is defined in Equation \eqref{eq:metric_loss}. This combines the intra-class loss and inter-class loss for all the classes.
\begin{equation} \label{eq:metric_loss}
\mathcal{L}_{\metric} = \sum_{k = 1}^n \bigg\{ \mathcal{L}_{\intra}(k) + \lambda \sum_{i \ne k} \mathcal{L}_{\inter}(k, i) \bigg\}
\end{equation}

where $\lambda$ is a pre-defined parameter to weight the importance of the intra- and inter-class losses. We set $\lambda$ to 0.1 by default.

Figure \ref{fig:metric} illustrates an example of a three-class feature representation in two dimensions. The left-hand figure shows the feature distribution trained with no metric learning. Obviously, the feature distance of the intra-class is large, sometimes even exceeding those of the inter-class distance near the decision boundary. However, the features trained by metric learning, shown in the right-hand figure, exhibit clear gaps between the inter-class predictions. This means the predictions with dropout are less likely to result in an inaccurate prediction and even reduce the variance of dropout prediction trials. The example shown in Figure \ref{fig:metric} has eight dropout predictions, three of which are classified to an inaccurate class when no metric learning is applied compared to only one inaccurate prediction with metric learning.


\subsection{Uncertainty Measurement} \label{section:model_uncertainty}
Bayesian models such as the Gaussian process \cite{rasmussen2004gaussian} provide a powerful tool to identify low-confidence regions of input space. 
Recently, Dropout \cite{JMLR:v15:srivastava14a}, which is used in deep neural networks, has been shown to serve as a Bayesian approximation to represent the model uncertainty in deep learning \cite{Gal:2016:DBA:3045390.3045502}. 
Based on this work, we propose a novel information-entropy-based dropout method to measure the model uncertainty in combination with metric learning for text classification.
Given an input data instance $\bm x^*$, we assume the corresponding output of our model is $y^*$. The output computed by our model incorporates a dropout mechanism in its evaluation mode, which means the activations of intermediate layers with Dropout are not reduced by a factor. 
When we repeat the process $k$ times, we obtain the output vector $\bm y^* = \{y_1^*, \dots, y_k^*\}$. Note that the outputs are not the same since the output here is generated by applying dropout after the feature representation layer in Figure \ref{fig:model}.

\begin{figure}[t]
	\centering
	\scalebox{0.65}{
		\includegraphics[trim=3.5cm 0.1cm 3cm 0.1cm, width=0.95\linewidth]{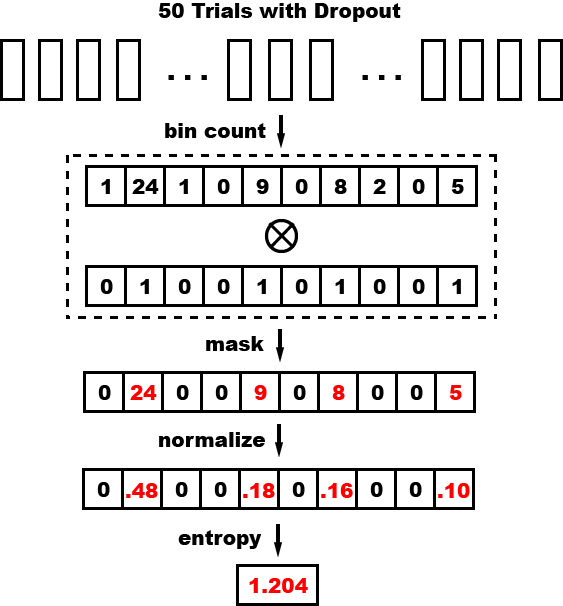}}	\caption{%
		\small Example of the dropout-entropy method.	}%
	\label{fig:uncertainty}
\end{figure}

Given the output $\bm y^*$ of $k$ trials with Dropout, our proposed uncertainty method has the following four steps, as shown in Figure \ref{fig:uncertainty}: 
(1) \textbf{Bin count}. We use bin count to calculate the frequency of each class. For example, if the class 2 appears 24 times in the dropout output vector $\bm y^*$, the bin count for class $2$ is $24$. 
(2) \textbf{Mask}. We use the mask step to avoid random noises in the frequency vector. In this step, we set the largest $m$ elements to have their original values and the remaining ones to zero. The value of $m$ is usually chosen to be $2/3$ of the total class number when the total classes are over 10; otherwise, we just skip the step.
(3) \textbf{Normalization}. We use the normalization step to calculate the probabilities of each class.
(4) \textbf{Information entropy}. The information entropy is calculated by $u = - \sum_{i=1}^{c} p_k(i) \log p_k(i)$,
where $p_k(i)$ represents the frequency probability of the $i$-th class in a total $k$ trials and $c$ is the number of classes. We use the entropy value as the uncertainty score here, in which the smaller the entropy value is, the more confident the model is in the output. Take the case in Figure \ref{fig:uncertainty} as an example. When the frequency of class 2 is 24, the entropy is $1.204$. If the output of the $50$ trials all belong to class $2$, the entropy becomes $0.401$, which means that the model is less uncertain about the predictive results. 

\begin{table*}[ht]
	\centering
	\small
	
	\scalebox{1.15}{
		\begin{tabularx}{0.85\textwidth}{c|c *{4}{Y}}
			\toprule
			& \multicolumn{5}{c}{\textbf{Uncertainty Ratio} (\textbf{Micro F1, Improved Ratio})} \\
			\cmidrule(lr){2-6} 
			& \textbf{0\%} & \textbf{10\%} & \textbf{20\%} & \textbf{30\%} & \textbf{40\%} \\
			\midrule
			
			\textbf{PL-Variance}		& 0.760 & 0.799(5.10\%) & 0.815(7.30\%) & 0.827(8.87\%)	& 0.840(10.52\%)\\	
			\textbf{Distance}			& 0.780 & 0.784(0.59\%) & 0.787(0.94\%) & 0.795(1.91\%)	& 0.800(2.58\%)\\	
			\textbf{NNGP}			& 0.637 & 0.659(3.44\%) & 0.670(5.04\%) & 0.678(6.44\%)	& 0.689(8.11\%)\\	
			\textbf{Dropout}			& 0.758 & 0.792(4.47\%) & 0.827(9.08\%) & 0.851(12.18\%)	& 0.879(15.93\%)\\
			\midrule
			\textbf{Dropout + Metric}	& 0.781 & 0.823(5.38\%) & 0.863(10.53\%) & 0.892(14.31\%)	& 0.921(18.05\%)\\	
			\midrule
			\textbf{DE}					& 0.760 & 0.807(6.25\%) & 0.849(11.73\%) & 0.888(16.79\%)	& 0.917(20.70\%)\\	
			\textbf{DE + Metric}		& 0.781 & \textbf{0.835}(\textbf{6.93}\%) & \textbf{0.878}(\textbf{12.47\%}) & \textbf{0.918}(\textbf{17.62\%})	& \textbf{0.944}(\textbf{20.92\%})\\	
		\end{tabularx}}

		\scalebox{1.15}{
		\begin{tabularx}{0.85\textwidth}{c|c *{4}{Y}}
			\toprule
			& \multicolumn{5}{c}{\textbf{Uncertainty Ratio} (\textbf{Macro F1, Improved Ratio})} \\
			\cmidrule(lr){2-6} 
			& \textbf{0\%} & \textbf{10\%} & \textbf{20\%} & \textbf{30\%} & \textbf{40\%} \\
			\midrule
			
			\textbf{PL-Variance}		& 0.751 & 0.789(5.05\%) & 0.806(7.24\%) & 0.818(8.87\%)	& 0.830(10.49\%)\\	
			\textbf{Distance}			& 0.773 & 0.777(0.48\%) & 0.779(0.81\%) & 0.786(1.76\%)	& 0.789(2.15\%)\\	
			\textbf{NNGP}			& 0.624 & 0.647(3.56\%) & 0.657(5.27\%) & 0.665(6.54\%)	& 0.675(8.13\%)\\	
			\textbf{Dropout}			& 0.749 & 0.781(4.22\%) & 0.813(8.46\%) & 0.833(11.10\%)	& 0.860(14.74\%)\\
			\midrule
			\textbf{Dropout + Metric}	& 0.773 & 0.816(5.47\%) & 0.853(10.33\%) & 0.878(13.59\%)	& 0.906(17.14\%)\\	
			\midrule
			\textbf{DE}					& 0.752 & 0.796(5.96\%) & 0.835(11.05\%) & 0.872(16.04\%)	& 0.900(19.70\%)\\	
			\textbf{DE + Metric}		& 0.774 & \textbf{0.826}(\textbf{6.70}\%) & \textbf{0.866}(\textbf{11.97\%}) & \textbf{0.904}(\textbf{16.87\%})	& \textbf{0.929}(\textbf{20.02\%})\\	
			\bottomrule
		\end{tabularx}}
	\vspace{0.2em}
	\caption{Uncertainty Scores for the 20 NewsGroup Dataset (20 Categories)}
	\label{table:20news_result}
\end{table*}

\section{Experiment} \label{section:experiment}
In this section, the performance of the proposed model uncertainty approach is evaluated on multiple real-world document classification data sets. After an introduction of the experiment settings in Section \ref{section:experiment_setup}, we compare the performance achieved by the proposed method against those of existing state-of-the-art methods, along with an analysis of the parameter settings and metric learning in Section \ref{section:performance}. Due to space limitation, the detailed experiment results on different sequence models can be accessed in the full version here\footnote{\url{https://xuczhang.github.io/papers/naacl19_uncertainty_full.pdf}}. The source code can be downloaded here\footnote{\url{https://github.com/xuczhang/UncertainDC}}.

\subsection{Experimental Setup} \label{section:experiment_setup}
In our experiments, all word vectors are initialized by pre-trained Glove \cite{pennington2014glove} word vectors, by default. The word embedding vectors are pre-trained in Wikipedia 2014 with a word vector dimension of 200. 
We trained all the DNN-based models with a batch size of 32 samples with a momentum of 0.9 and an initial learning rate of 0.001 using the Adam \cite{kingma2014adam} optimization algorithm.

\subsubsection{Datasets and Labels}
We conducted experiments on three publicly available datasets: 
1) \textbf{20 Newsgroups}\footnote{\url{http://qwone.com/~jason/20Newsgroups/}} \cite{Lang95}: The data set is a collection of 20,000 documents, partitioned evenly across 20 different news groups; 
2) \textbf{IMDb Reviews} \cite{maas-EtAl:2011:ACL-HLT2011}: The data set contains 50,000 popular movie reviews with binary positive or negative labels from the IMDb website; and
3) \textbf{Amazon Reviews} \cite{mcauley2013hidden}: The dataset is a collection of reviews from Amazon spanning the time period from May 1996 to July 2013. We used review data from the Sports and outdoors category, with 272,630 data samples and rating labels from 1 to 5. 

For all three data sets, we randomly selected 70\% of the data samples as the training set, 10\% as the validation set and 20\% as the test set.

\begin{table*}[h]
	\centering
	\small
	
	\scalebox{1.15}{
		\begin{tabularx}{0.85\textwidth}{c|c *{4}{Y}}
			\toprule
			& \multicolumn{5}{c}{\textbf{Uncertainty Ratio} (\textbf{Accuracy, Improved Ratio})} \\
			\cmidrule(lr){2-6} 
			& \textbf{0\%} & \textbf{10\%} & \textbf{20\%} & \textbf{30\%} & \textbf{40\%} \\
			\midrule
			
			\textbf{PL-Variance}		& 0.878 & 0.911(3.69\%) & 0.937(6.70\%) & 0.955(8.71\%)	& 0.970(\textbf{10.42\%})\\	
			\textbf{Distance}			& 0.884 & 0.893(0.95\%) & 0.892(0.91\%) & 0.893(1.04\%)	& 0.895(1.24\%)\\	
			\textbf{Dropout}			& 0.880 & 0.912(3.72\%) & 0.936(6.43\%) & 0.957(8.75\%)	& 0.969(10.20\%)\\
			\midrule
			\textbf{Dropout + Metric}	& 0.884 & 0.917(3.73\%) & \textbf{0.944}(6.78\%) & \textbf{0.961}(8.70\%)	& \textbf{0.973}(10.11\%)\\	
			\midrule
			\textbf{DE}					& 0.878 & 0.911(3.70\%) & 0.937(6.71\%) & 0.956(8.83\%)	& 0.969(10.33\%)\\	
			\textbf{DE + Metric}		& 0.883 & \textbf{0.918}(\textbf{3.91\%}) & \textbf{0.944}(\textbf{6.87\%}) & \textbf{0.961}(\textbf{8.78\%})	& \textbf{0.973}(10.20\%)\\

		\end{tabularx}}

		\scalebox{1.15}{
		\begin{tabularx}{0.85\textwidth}{c|c *{4}{Y}}
			\toprule
			& \multicolumn{5}{c}{\textbf{Uncertainty Ratio} (\textbf{F1 Score, Improved Ratio})} \\
			\cmidrule(lr){2-6} 
			& \textbf{0\%} & \textbf{10\%} & \textbf{20\%} & \textbf{30\%} & \textbf{40\%} \\
			\midrule
			
			\textbf{PL-Variance}		& 0.880 & 0.913(3.68\%) & 0.939(6.67\%) & 0.956(8.65\%)	& 0.971(\textbf{10.34\%})\\	
			\textbf{Distance}			& 0.885 & 0.894(1.07\%) & 0.898(1.42\%) & 0.901(1.84\%)	& 0.904(2.13\%)\\	
			\textbf{Dropout}			& 0.881 & 0.914(3.70\%) & 0.938(6.41\%) & 0.958(8.67\%)	& 0.971(10.13\%)\\
			\midrule
			\textbf{Dropout + Metric}	& 0.885 & 0.917(3.70\%) & \textbf{0.944}(6.74\%) & \textbf{0.961}(8.67\%)	& \textbf{0.974}(10.06\%)\\	
			\midrule
			\textbf{DE}					& 0.880 & 0.913(3.67\%) & 0.939(6.67\%) & 0.957(8.77\%)	& 0.970(10.25\%)\\	
			\textbf{DE + Metric}		& 0.884 & \textbf{0.918}(\textbf{3.88\%}) & \textbf{0.944}(\textbf{6.83\%}) & \textbf{0.961}(\textbf{8.73\%})	& \textbf{0.974}(10.14\%)\\
			\bottomrule
		\end{tabularx}}
	\vspace{0.2em}
	\caption{Uncertainty Scores for the IMDb Dataset (2 Categories)}
	\label{table:imdb_result}
\end{table*}

\begin{table*}[h]
	\centering
	\small
	
	\scalebox{1.15}{
		\begin{tabularx}{0.85\textwidth}{c|c *{4}{Y}}
			\toprule
			& \multicolumn{5}{c}{\textbf{Uncertainty Ratio} (\textbf{Accuracy, Improved Ratio})} \\
			\cmidrule(lr){2-6} 
			& \textbf{0\%} & \textbf{10\%} & \textbf{20\%} & \textbf{30\%} & \textbf{40\%} \\
			\midrule
			
			\textbf{PL-Variance}		& 0.700 & 0.738(5.43\%) & 0.764(9.14\%) & 0.784(1.20\%)	& 0.801(14.4\%)\\	
			\textbf{Distance}			& 0.697 & 0.699(0.29\%) & 0.702(0.72\%) & 0.704(1.00\%)	& 0.705(1.15\%)\\	
			\textbf{Dropout}			& 0.700 & 0.735(5.00\%) & 0.764(9.14\%) & 0.800(14.29\%)	& 0.831(18.71\%)\\
			\midrule
			\textbf{Dropout + Metric}	& 0.710 & 0.746(5.07\%) & 0.779(9.72\%) & 0.815(14.79\%)	& 0.847(19.30\%)\\	
			\midrule
			\textbf{DE}					& 0.700 & 0.739(\textbf{5.57\%}) & 0.773(10.43\%) & 0.806(15.14\%)	& 0.836(19.43\%)\\	
			\textbf{DE + Metric}		& 0.724 & \textbf{0.764}(5.52\%) & \textbf{0.800}(\textbf{10.50\%}) & \textbf{0.834}(\textbf{15.19\%})	& \textbf{0.866}(\textbf{19.61\%})\\

			\bottomrule
		\end{tabularx}}
	\vspace{0.2em}
	\caption{Uncertainty Scores for the Amazon Dataset (5 Categories)}
	\label{table:amazon_result}
\end{table*}

\begin{figure}[tb]
	\centering
	\scalebox{0.6}{
		\includegraphics[trim=3cm 0cm 4cm 0.1cm, width=0.95\linewidth]{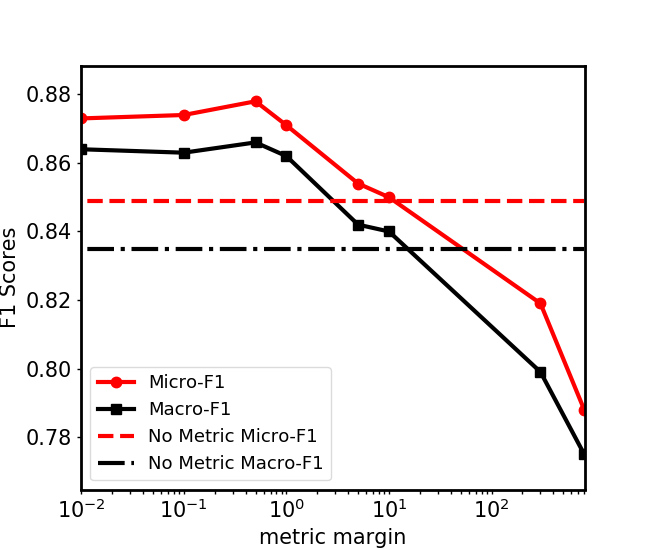}}	\caption{%
		\small Prediction performance for different metric margin settings.}%
	\label{fig:metric_margin}
\end{figure}

\subsubsection{Evaluation Metrics}
In order to answer the question "\textit{What percentage of data should be transferred to domain experts to achieve an overall accuracy rate above 90\%?}", we measure the classification performance in terms of various uncertainty ratios. Specifically, assuming the entire testing set $S$ has size $n$ and an uncertainty ratio $r$, we can remove the most uncertain samples $S_r$ from $S$ based on the uncertainty ratio $r$, where the size of the uncertainty set $S_r$ is $r\cdot n$. We assume the uncertain samples $S_r$ handed to domain experts achieve 100\% accuracy. If the uncertainty ratio $r$ equals to $0$, the model performs without uncertainty measurement concerns.

For the binary classification task, we use the accuracy and F1-score to measure the classification performance based on the testing set $S \setminus S_r$ for different uncertainty ratios $r$. Similarly, for multi-class tasks, we use the micro-F1 and macro-F1 scores utilizing the same settings as for the binary classification.

\begin{table*}[t]
	\centering
	\small

	\scalebox{1.15}{
		\begin{tabularx}{0.85\textwidth}{c|cc *{4}{Y}}
			\toprule
			& \multicolumn{6}{c}{\textbf{Uncertainty Ratio} (\textbf{Micro F1, Improved Ratio})} \\
			\cmidrule(lr){2-7} 
			&& \textbf{0\%} & \textbf{10\%} & \textbf{20\%} & \textbf{30\%} & \textbf{40\%} \\
			\midrule

			\multirow{2}{*}{\textbf{Random}} &\textbf{DE}					& 0.659 & 0.702(6.47\%) & 0.748(13.46\%) & 0.792(20.14\%)	& 0.831(26.03\%)\\	
			&\textbf{DE + Metric}		& 0.660 & \textbf{0.705}(\textbf{6.85\%}) & \textbf{0.752}(\textbf{13.92\%}) & \textbf{0.802}(\textbf{21.57\%})	& \textbf{0.845}(\textbf{28.04\%})\\
			\midrule
			\multirow{2}{*}{\textbf{Glove}} &\textbf{DE}					& 0.760 & 0.807(6.25\%) & 0.849(11.73\%) & 0.888(16.79\%)	& 0.917(20.70\%)\\	
			&\textbf{DE + Metric}		& 0.781 & \textbf{0.835}(\textbf{6.93}\%) & \textbf{0.878}(\textbf{12.47\%}) & \textbf{0.918}(\textbf{17.62\%})	& \textbf{0.944}(\textbf{20.92\%})\\	
			\bottomrule
		\end{tabularx}}
	\vspace{0.2em}
	\caption{Embedding vs. No Pre-trained Embedding}
	\label{table:embedding}
\end{table*}

\subsubsection{Comparison Methods}
The following methods are included in the performance comparison: 
1) Penultimate Layer Variance (PL-Variance). Activations before the softmax layer in a deep neural network always reveal the uncertainty of the prediction \cite{zaragoza1998confidence}. As a baseline method, we use the variance of the output of a fully connected layer in Figure \ref{fig:model} as the uncertainty weight.
2) Deep Neural Networks as Gaussian Processes (NNGP) \cite{lee2017deep}. This approach applies a Gaussian process to perform a Bayesian inference for deep neural networks, with a computationally efficient pipeline being used to compute the covariance function of the Gaussian process. The default parameter settings in the source code\footnote{https://github.com/brain-research/nngp} were applied in our experiments. 
3) Distance-based Confidence (Distance)\cite{mandelbaum2017distance}. This method assigns confidence scores based on the data embedding compared to the training data. We set its nearest neighbor parameter $k=10$.
4) Dropout \cite{Gal:2016:DBA:3045390.3045502}. Here, dropout training in DNNs is treated as an approximation of Bayesian inference in deep Gaussian processes. We set the sample number $T$ as 100 in our experiments.
5) Dropout + Metric. In order to validate the effectiveness of our metric learning, we applied our proposed metric learning method to the Dropout method. The metric margin $m$ and coefficient $\lambda$ were set as $0.5$ and $0.1$, respectively. 
6) Our proposed method. We evaluate our proposed method in two different settings, Dropout-Entropy alone (DE) and Dropout-Entropy with metric learning (DE + Metric). Here, we set the sample number $T=100$, coefficient $\lambda =0.1$ and the metric margin may vary from different data sets. 

\subsection{Experimental Results} \label{section:performance}
This subsection presents the results of the uncertainty performance comparison and the analysis of the metric learning and parameter settings.

\subsubsection{Uncertainty Results}
Table \ref{table:20news_result} shows the Micro-F1 and Macro-F1 scores for ratios of uncertain predictions eliminated ranging from 10\% to 40\% for the 20NewsGroup data set. To demonstrate its effect, metric learning was also applied to the baseline method Dropout, and our proposed method DE. The improvement ratio compared to the results with no uncertainty elimination, shown in the $0\%$ column, are presented after the F1 scores. 
Based on these result, we can conclude that: 
1) Our proposed method, DE+Metric, significantly improves both the Micro- and Macro-F1 scores when a portion of uncertain predictions are eliminated. For example, the Micro-F1 improves from 0.78 to 0.92 when 30\% of the uncertain predictions are eliminated.
2) Comparing the results obtained by DE and DE+Metric, metric learning significantly improves the results obtained for different uncertainty ratio settings. Similar results can be observed when comparing the Dropout and Dropout+Metric. For example, the Micro-F1 scores for Dropout+Metric are around 5\% better than the Dropout method alone, boosting them from 0.851 to 0.892, with a 30\% uncertainty ratio.  
3) The DE method outperforms all the other methods when metric learning is not applied. Specifically, DE is around 4\% better than the Dropout method in terms of the Micro-F1 score.

\begin{figure*}[ht]
	\centering
	\subfigure[No Metric Learning]{%
		\label{fig:embed1}
		\includegraphics[trim=0.3cm 0.1cm 0.6cm 0.5cm,width=0.48\linewidth]{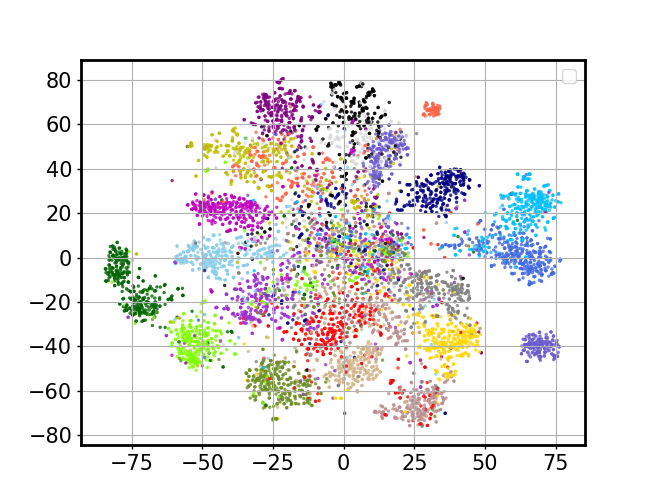}
	} %
	\subfigure[Metric Margin $m$ = 10]{%
		\label{fig:embed4}
		\includegraphics[trim=0.3cm 0.1cm 0.6cm 0.5cm,width=0.48\linewidth]{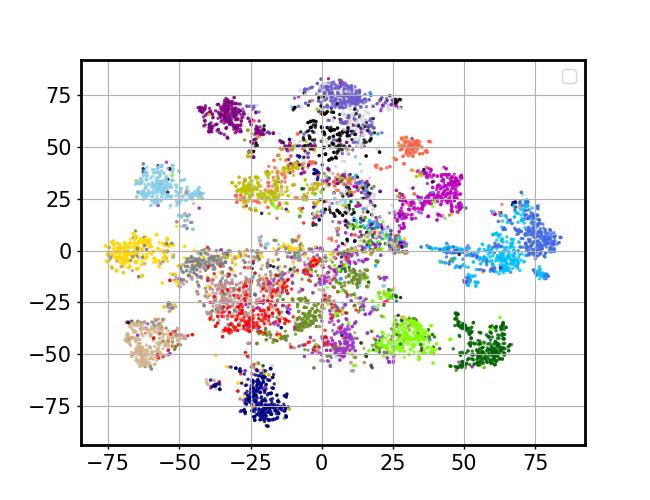}
	} %
	
	\caption{%
		\small Feature Visualization of 20 NewsGroup Testing Data Set in Two Dimensions by t-SNE Algorithm.
	}%
	\label{fig:embedding}
\end{figure*}

The results for IMDb and Amazon data sets are presented in Table \ref{table:imdb_result} and Table \ref{table:amazon_result}. When comparing our proposed model's performance across three data sets, we found that the greater improvements are achieved on multi- instead of binary-class classification data sets. One possible explanation is that a comparatively large portion of multi-class features are close to the decision boundary in the feature space. Through the metric learning strategy of minimizing intra-class distance while maxmizing the inter-class instances, the feature distance between the inter-class predictions is enlarged and the quality of embeddings is greatly enhanced.

\subsubsection{Analysis of Metric Learning}
The impact of metric learning on feature representation is analyzed in this section. Figure \ref{fig:embedding} shows the 300-dimension feature representations for the 20 NewsGroup testing data set, with Figure \ref{fig:embed1} presenting the features trained without metric learning and Figure \ref{fig:embed4} that trained by metric learning with a margin parameter $m$=10. We used the t-SNE algorithm \cite{maaten2008visualizing} to visualize the high dimensional features in the form of two dimensional images. From the results, we can clearly see that the distances between the inter-classes are significantly enlarged compared to the features trained without metric learning shown in Figure \ref{fig:embed1}. This enlarged inter-class spacing means that dropout-based uncertainty methods have smaller prediction variances in case their dropout prediction trials are accurate.

\subsubsection{Parameter Analysis}
The impact of the metric margin and word embeddings are discussed in this section.

\textit{Metric Margin.}
Figure \ref{fig:metric_margin} shows the impact of metric margin parameters, ranging from 0 to 800 on the 20 NewsGroup data set with a 20\% uncertainty ratio. From the results, we can conclude that: (1) The prediction performance is not sensitive to the point at which the metric margin parameter is set as long as its value is not extremely large. (2) Compared to the model trained with no metric learning, our methods consistently achieve better performance when the metric margin is set no larger than 10. When the metric margin is too large, however, the prediction cross-entropy loss is hard to minimize and thus dampens the overall prediction performance. (3) The results of Macro-F1 are similar to Micro-F1 with relatively small scores. 

\textit{Impact of Word Embedding.} We also analyzed the impact of our proposed methods on different word embedding initialization methods, including random and pre-trained Glove word vectors in 200 dimensions. Table \ref{table:embedding} shows the results of Micro-F1 for the different uncertainty ratios. We can observe that: 1) The performance of Glove-based methods are around 15\% better than that of the randomly initialized methods for different uncertainty ratios. 2) Metric learning based on a Glove initialization generally outperforms a random initialization. For instance, the F1 score of Glove rises by 0.29 when the uncertainty ratio is 20\%, while for a random method it only increases by 0.04.

\section{Conclusion} \label{section:conclusion}
In this paper, a DNN-based model is proposed to address the uncertainty mitigation problem in the presence of human involvement in a text classification task. To achieve this, we proposed a dropout-entropy uncertainty measurement method with the metric learning for the feature representation. Extensive experiments on real-world data sets confirmed that our proposed approach dramatically outperforms competing methods, exhibiting a significant improvement in accuracy when a relatively small portion of the uncertainty predictions are handed over to domain experts.
   
%
%

\newpage
\bibliography{naaclhlt2019}

\begin{thebibliography}{33}
\expandafter\ifx\csname natexlab\endcsname\relax\def\natexlab#1{#1}\fi

\bibitem[{Bahdanau et~al.(2014)Bahdanau, Cho, and Bengio}]{bahdanau2014neural}
Dzmitry Bahdanau, Kyunghyun Cho, and Yoshua Bengio. 2014.
\newblock Neural machine translation by jointly learning to align and
  translate.
\newblock \emph{arXiv preprint arXiv:1409.0473}.

\bibitem[{Denker and LeCun(1990)}]{Denker:1990:TNO:2986766.2986882}
John~S. Denker and Yann LeCun. 1990.
\newblock \href {http://dl.acm.org/citation.cfm?id=2986766.2986882}
  {Transforming neural-net output levels to probability distributions}.
\newblock In \emph{Proceedings of the 3rd International Conference on Neural
  Information Processing Systems}, NIPS'90, pages 853--859, San Francisco, CA,
  USA. Morgan Kaufmann Publishers Inc.

\bibitem[{Gal and Ghahramani(2016)}]{Gal:2016:DBA:3045390.3045502}
Yarin Gal and Zoubin Ghahramani. 2016.
\newblock \href {http://dl.acm.org/citation.cfm?id=3045390.3045502} {Dropout as
  a bayesian approximation: Representing model uncertainty in deep learning}.
\newblock In \emph{Proceedings of the 33rd International Conference on
  International Conference on Machine Learning - Volume 48}, ICML'16, pages
  1050--1059. JMLR.org.

\bibitem[{Gong et~al.(2013)Gong, Liang, Shi, Ma, and Ma}]{gong2013fuzzy}
Maoguo Gong, Yan Liang, Jiao Shi, Wenping Ma, and Jingjing Ma. 2013.
\newblock Fuzzy c-means clustering with local information and kernel metric for
  image segmentation.
\newblock \emph{IEEE Transactions on Image Processing}, 22(2):573--584.

\bibitem[{Goodfellow et~al.(2014)Goodfellow, Shlens, and
  Szegedy}]{goodfellow2014explaining}
Ian~J Goodfellow, Jonathon Shlens, and Christian Szegedy. 2014.
\newblock Explaining and harnessing adversarial examples.
\newblock \emph{arXiv preprint arXiv:1412.6572}.

\bibitem[{Guillaumin et~al.(2009)Guillaumin, Verbeek, and
  Schmid}]{guillaumin2009you}
Matthieu Guillaumin, Jakob Verbeek, and Cordelia Schmid. 2009.
\newblock Is that you? metric learning approaches for face identification.
\newblock In \emph{Computer Vision, 2009 IEEE 12th international conference
  on}, pages 498--505. IEEE.

\bibitem[{Hern{\'a}ndez-Lobato and Adams(2015)}]{hernandez2015probabilistic}
Jos{\'e}~Miguel Hern{\'a}ndez-Lobato and Ryan Adams. 2015.
\newblock Probabilistic backpropagation for scalable learning of bayesian
  neural networks.
\newblock In \emph{International Conference on Machine Learning}, pages
  1861--1869.

\bibitem[{Hsieh et~al.(2017)Hsieh, Yang, Cui, Lin, Belongie, and
  Estrin}]{hsieh2017collaborative}
Cheng-Kang Hsieh, Longqi Yang, Yin Cui, Tsung-Yi Lin, Serge Belongie, and
  Deborah Estrin. 2017.
\newblock Collaborative metric learning.
\newblock In \emph{Proceedings of the 26th International Conference on World
  Wide Web}, pages 193--201. International World Wide Web Conferences Steering
  Committee.

\bibitem[{Iyyer et~al.(2014)Iyyer, Boyd-Graber, Claudino, Socher, and
  Daum{\'e}~III}]{iyyer2014neural}
Mohit Iyyer, Jordan Boyd-Graber, Leonardo Claudino, Richard Socher, and Hal
  Daum{\'e}~III. 2014.
\newblock A neural network for factoid question answering over paragraphs.
\newblock In \emph{Proceedings of the 2014 Conference on Empirical Methods in
  Natural Language Processing (EMNLP)}, pages 633--644.

\bibitem[{Kendall and Gal(2017)}]{kendall2017uncertainties}
Alex Kendall and Yarin Gal. 2017.
\newblock What uncertainties do we need in bayesian deep learning for computer
  vision?
\newblock In \emph{Advances in neural information processing systems}, pages
  5574--5584.

\bibitem[{Kingma and Ba(2014)}]{kingma2014adam}
Diederik~P Kingma and Jimmy Ba. 2014.
\newblock Adam: A method for stochastic optimization.
\newblock \emph{arXiv preprint arXiv:1412.6980}.

\bibitem[{Lakshminarayanan et~al.(2017)Lakshminarayanan, Pritzel, and
  Blundell}]{lakshminarayanan2017simple}
Balaji Lakshminarayanan, Alexander Pritzel, and Charles Blundell. 2017.
\newblock Simple and scalable predictive uncertainty estimation using deep
  ensembles.
\newblock In \emph{Advances in Neural Information Processing Systems}, pages
  6405--6416.

\bibitem[{Lang(1995)}]{Lang95}
Ken Lang. 1995.
\newblock Newsweeder: Learning to filter netnews.
\newblock In \emph{Proceedings of the Twelfth International Conference on
  Machine Learning}, pages 331--339.

\bibitem[{Lee et~al.(2017)Lee, Bahri, Novak, Schoenholz, Pennington, and
  Sohl-Dickstein}]{lee2017deep}
Jaehoon Lee, Yasaman Bahri, Roman Novak, Samuel~S Schoenholz, Jeffrey
  Pennington, and Jascha Sohl-Dickstein. 2017.
\newblock Deep neural networks as gaussian processes.
\newblock \emph{arXiv preprint arXiv:1711.00165}.

\bibitem[{Maas et~al.(2011)Maas, Daly, Pham, Huang, Ng, and
  Potts}]{maas-EtAl:2011:ACL-HLT2011}
Andrew~L. Maas, Raymond~E. Daly, Peter~T. Pham, Dan Huang, Andrew~Y. Ng, and
  Christopher Potts. 2011.
\newblock \href {http://www.aclweb.org/anthology/P11-1015} {Learning word
  vectors for sentiment analysis}.
\newblock In \emph{Proceedings of the 49th Annual Meeting of the Association
  for Computational Linguistics: Human Language Technologies}, pages 142--150,
  Portland, Oregon, USA. Association for Computational Linguistics.

\bibitem[{Maaten and Hinton(2008)}]{maaten2008visualizing}
Laurens van~der Maaten and Geoffrey Hinton. 2008.
\newblock Visualizing data using t-sne.
\newblock \emph{Journal of machine learning research}, 9(Nov):2579--2605.

\bibitem[{Mandelbaum and Weinshall(2017)}]{mandelbaum2017distance}
Amit Mandelbaum and Daphna Weinshall. 2017.
\newblock Distance-based confidence score for neural network classifiers.
\newblock \emph{arXiv preprint arXiv:1709.09844}.

\bibitem[{McAllister et~al.(2017)McAllister, Gal, Kendall, Van Der~Wilk, Shah,
  Cipolla, and Weller}]{mcallister2017concrete}
Rowan McAllister, Yarin Gal, Alex Kendall, Mark Van Der~Wilk, Amar Shah,
  Roberto Cipolla, and Adrian~Vivian Weller. 2017.
\newblock Concrete problems for autonomous vehicle safety: Advantages of
  bayesian deep learning.
\newblock International Joint Conferences on Artificial Intelligence, Inc.

\bibitem[{McAuley and Leskovec(2013)}]{mcauley2013hidden}
Julian McAuley and Jure Leskovec. 2013.
\newblock Hidden factors and hidden topics: understanding rating dimensions
  with review text.
\newblock In \emph{Proceedings of the 7th ACM conference on Recommender
  systems}, pages 165--172. ACM.

\bibitem[{Mithe et~al.(2013)Mithe, Indalkar, and Divekar}]{mithe2013optical}
Ravina Mithe, Supriya Indalkar, and Nilam Divekar. 2013.
\newblock Optical character recognition.
\newblock \emph{International journal of recent technology and engineering
  (IJRTE)}, 2(1):72--75.

\bibitem[{Parkhi et~al.(2015)Parkhi, Vedaldi, Zisserman
  et~al.}]{parkhi2015deep}
Omkar~M Parkhi, Andrea Vedaldi, Andrew Zisserman, et~al. 2015.
\newblock Deep face recognition.
\newblock In \emph{BMVC}, volume~1, page~6.

\bibitem[{Pennington et~al.(2014)Pennington, Socher, and
  Manning}]{pennington2014glove}
Jeffrey Pennington, Richard Socher, and Christopher~D. Manning. 2014.
\newblock \href {http://www.aclweb.org/anthology/D14-1162} {Glove: Global
  vectors for word representation}.
\newblock In \emph{Empirical Methods in Natural Language Processing (EMNLP)},
  pages 1532--1543.

\bibitem[{Rasmussen(2004)}]{rasmussen2004gaussian}
Carl~Edward Rasmussen. 2004.
\newblock Gaussian processes in machine learning.
\newblock In \emph{Advanced lectures on machine learning}, pages 63--71.
  Springer.

\bibitem[{Shafer and Vovk(2008)}]{shafer2008tutorial}
Glenn Shafer and Vladimir Vovk. 2008.
\newblock A tutorial on conformal prediction.
\newblock \emph{Journal of Machine Learning Research}, 9(Mar):371--421.

\bibitem[{Srivastava et~al.(2014)Srivastava, Hinton, Krizhevsky, Sutskever, and
  Salakhutdinov}]{JMLR:v15:srivastava14a}
Nitish Srivastava, Geoffrey Hinton, Alex Krizhevsky, Ilya Sutskever, and Ruslan
  Salakhutdinov. 2014.
\newblock \href {http://jmlr.org/papers/v15/srivastava14a.html} {Dropout: A
  simple way to prevent neural networks from overfitting}.
\newblock \emph{Journal of Machine Learning Research}, 15:1929--1958.

\bibitem[{Vovk et~al.(1999)Vovk, Gammerman, and Saunders}]{vovk1999machine}
Volodya Vovk, Alexander Gammerman, and Craig Saunders. 1999.
\newblock Machine-learning applications of algorithmic randomness.

\bibitem[{Wang and Yeung(2016)}]{wang2016towards}
Hao Wang and Dit-Yan Yeung. 2016.
\newblock Towards bayesian deep learning: A survey.
\newblock \emph{arXiv preprint arXiv:1604.01662}.

\bibitem[{Weinberger et~al.(2006)Weinberger, Blitzer, and
  Saul}]{weinberger2006distance}
Kilian~Q Weinberger, John Blitzer, and Lawrence~K Saul. 2006.
\newblock Distance metric learning for large margin nearest neighbor
  classification.
\newblock In \emph{Advances in neural information processing systems}, pages
  1473--1480.

\bibitem[{van~der Westhuizen and Lasenby(2017)}]{van2017bayesian}
Jos van~der Westhuizen and Joan Lasenby. 2017.
\newblock Bayesian lstms in medicine.
\newblock \emph{arXiv preprint arXiv:1706.01242}.

\bibitem[{Xing et~al.(2003)Xing, Jordan, Russell, and Ng}]{xing2003distance}
Eric~P Xing, Michael~I Jordan, Stuart~J Russell, and Andrew~Y Ng. 2003.
\newblock Distance metric learning with application to clustering with
  side-information.
\newblock In \emph{Advances in neural information processing systems}, pages
  521--528.

\bibitem[{Xu et~al.(2012)Xu, Chen, Weinberger, and
  Sha}]{Xu:2012:SDM:2396761.2398536}
Zhixiang~(Eddie) Xu, Minmin Chen, Kilian~Q. Weinberger, and Fei Sha. 2012.
\newblock \href {https://doi.org/10.1145/2396761.2398536} {From sbow to dcot
  marginalized encoders for text representation}.
\newblock In \emph{Proceedings of the 21st ACM International Conference on
  Information and Knowledge Management}, CIKM '12, pages 1879--1884, New York,
  NY, USA. ACM.

\bibitem[{Yang et~al.(2016)Yang, Yang, Dyer, He, Smola, and
  Hovy}]{yang2016hierarchical}
Zichao Yang, Diyi Yang, Chris Dyer, Xiaodong He, Alex Smola, and Eduard Hovy.
  2016.
\newblock Hierarchical attention networks for document classification.
\newblock In \emph{Proceedings of the 2016 Conference of the North American
  Chapter of the Association for Computational Linguistics: Human Language
  Technologies}, pages 1480--1489.

\bibitem[{Zaragoza and d'Alche Buc(1998)}]{zaragoza1998confidence}
Hugo Zaragoza and Florence d'Alche Buc. 1998.
\newblock Confidence measures for neural network classifiers.
\newblock In \emph{Proceedings of the Seventh Int. Conf. Information Processing
  and Management of Uncertainty in Knowlegde Based Systems}.

\end{thebibliography}
\bibliographystyle{acl_natbib}

\newpage
\clearpage
\appendix
\section{Additional Experiments}

\label{sec:appendix}
In this section we present the results of additional experiments. To understand to what extend the proposed method is able to improve sequence model, we evaluated the performance of Bi-LSTM model on the 20 NewsGroup data set. Table \ref{table:20news_lstm_result} shows the accuracy scores of Bi-LSTM method for ratios of uncertain predictions eliminated ranging from 10\% to 40\%. To demonstrate the effect of metric learning, this was also applied to the baseline methods, Dropout, and our proposed method DE. Based on these results, we can conclude that: 1) Our proposed method, DE+Metric, markedly improves the accuracy when uncertain predictions are eliminated. For example, the accuracy improved from 0.694\% to 0.792\% when 30\% of the uncertain predictions are eliminated. 2) Comparing the results obtained by DE and DE+Metric, metric learning significantly improves the accuracy and F1 scores obtained for different uncertainty ratio settings. Similar results can be observed when comparing the Dropout and Dropout+Metric methods. For example, the F1 scores for Dropout+Metric are around 2\% better than the Dropout method alone. 3) The DE method outperforms all the other methods when metric learning is not applied. However, the performance improvement achieved by different uncertainty ratios in Bi-LSTM is relatively small comparing to that of CNN model.

\begin{figure*}[ht]
	\centering
	\subfigure[Accuracy]{%
		\label{fig:acc}
		\includegraphics[trim=0.3cm 0.1cm 0.6cm 0.5cm,width=0.48\linewidth]{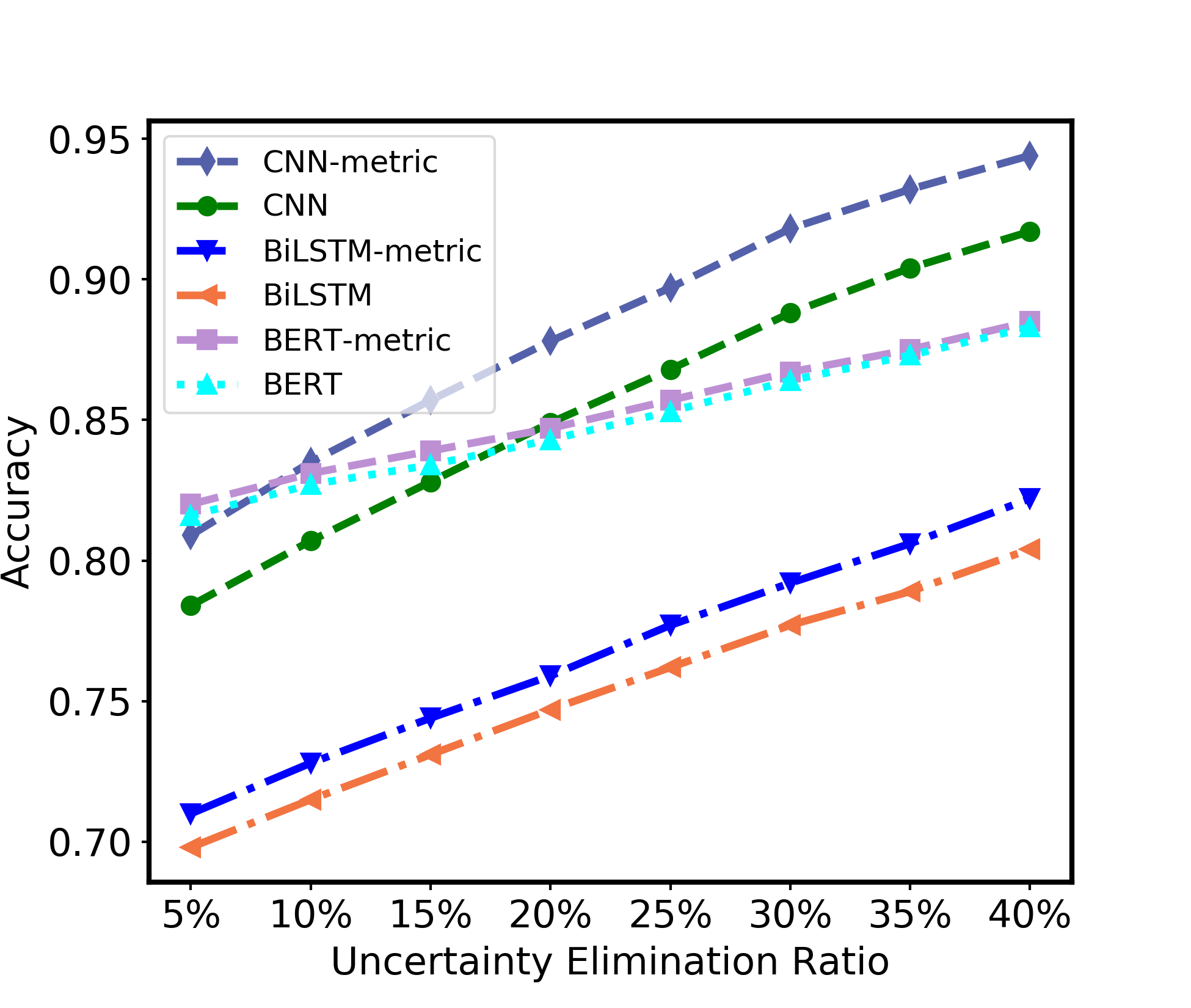}
	} %
	\subfigure[Improved Accuracy]{%
		\label{fig:improved_acc}
		\includegraphics[trim=0.3cm 0.1cm 0.6cm 0.5cm,width=0.48\linewidth]{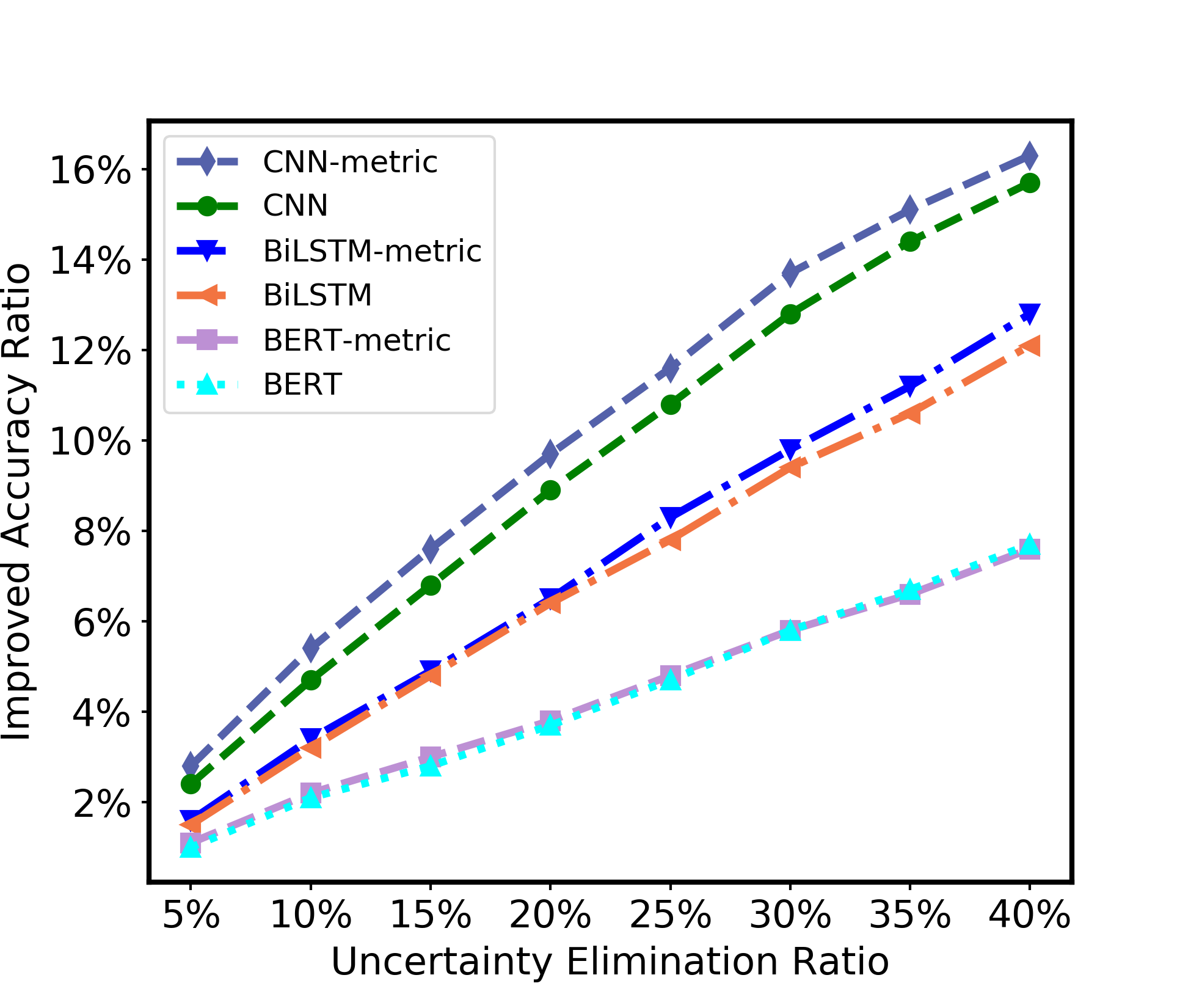}
	} %
	
	\caption{%
		\small Prediction accuracy and improved accuracy ratio of DNN models for 20 NewsGroup testing data set.
	}%
	\label{fig:compare}
\end{figure*}

\begin{table*}[h]
	\centering
	\small
	
	\scalebox{1.15}{
		\begin{tabularx}{0.85\textwidth}{c|c *{4}{Y}}
			\toprule
			& \multicolumn{5}{c}{\textbf{Uncertainty Ratio} (\textbf{Accuracy, Improved Ratio})} \\
			\cmidrule(lr){2-6} 
			& \textbf{0\%} & \textbf{10\%} & \textbf{20\%} & \textbf{30\%} & \textbf{40\%} \\
			\midrule
			
			\textbf{PL-Variance}		& 0.684 & 0.705(3.07\%) & 0.726(6.14\%) & 0.748(9.36\%)	& 0.771(12.72\%)\\	
			\textbf{Distance}			& 0.696 & 0.717(3.02\%) & 0.746(7.18\%) & 0.777(11.64\%)	& 0.806(15.80\%)\\	
			\textbf{Dropout}			& 0.683 & 0.717(\textbf{4.98\%}) & 0.749(\textbf{9.66\%}) & 0.779(14.06\%)	& 0.810(\textbf{18.59\%})\\
			\midrule
			\textbf{Dropout + Metric}	& 0.692 & 0.723(4.48\%) & 0.755(9.10\%) & 0.784(13.29\%)	& 0.813(17.49\%)\\	
			\midrule
			\textbf{DE}					& 0.684 & 0.716(4.61\%) & 0.749(9.37\%) & 0.781(13.98\%)	& 0.813(\textbf{18.59\%})\\	
			\textbf{DE + Metric}		& 0.694 & \textbf{0.728}(4.90\%) & \textbf{0.759}(9.37\%) & \textbf{0.792}(\textbf{14.12\%})	& \textbf{0.822}(18.44\%)\\	
		\end{tabularx}}

		\scalebox{1.15}{
		\begin{tabularx}{0.85\textwidth}{c|c *{4}{Y}}
			\toprule
			& \multicolumn{5}{c}{\textbf{Uncertainty Ratio} (\textbf{F1, Improved Ratio})} \\
			\cmidrule(lr){2-6} 
			& \textbf{0\%} & \textbf{10\%} & \textbf{20\%} & \textbf{30\%} & \textbf{40\%} \\
			\midrule
			
			\textbf{PL-Variance}		& 0.677 & 0.698(3.10\%) & 0.718(6.06\%) & 0.740(9.31\%)	& 0.763(12.70\%)\\	
			\textbf{Distance}			& 0.690 & 0.711(3.04\%) & 0.740(7.25\%) & 0.771(11.74\%)	& 0.801(16.09\%)\\	
			\textbf{Dropout}			& 0.675 & 0.711(\textbf{5.33\%}) & 0.744(\textbf{10.22\%}) & 0.775(\textbf{14.81\%})	& 0.807(19.56\%)\\
			\midrule
			\textbf{Dropout + Metric}	& 0.687 & 0.719(4.66\%) & 0.750(9.17\%) & 0.779(13.39\%)	& 0.809(17.76\%)\\	
			\midrule
			\textbf{DE}					&  0.677 & 0.710(4.87\%) & 0.744(9.90\%) & 0.777(14.77\%)	&  0.810(\textbf{19.65\%})\\	
			\textbf{DE + Metric}		& 0.689 & \textbf{0.723}(4.93\%) & \textbf{0.755}(9.58\%) & \textbf{0.788}(14.37\%)	& \textbf{0.819}(18.87\%)\\	
			\bottomrule
		\end{tabularx}}
	\vspace{0.2em}
	\caption{Uncertainty Scores for the 20 NewsGroup Dataset (20 Categories)[Bi-LSTM]}
	\label{table:20news_lstm_result}
\end{table*}

We further extended experiments on BERT, one of the state of the art pre-trained language models. The comparative results of three DNN-based models using the proposed DE and DE+Metric method are reported in Figure \ref{fig:compare}. Based on the results, we can observe that: 1) BERT performs the best in regards of accuracy with 5\% of uncertain predictions eliminated. 2) The accuracy achieved by CNN model exceeds BERT as the uncertainty ratio increases, and performs better than the others when the uncertainty ratio is high. 3) Bi-LSTM performs worst among the three models with a 0.822 accuracy when 40\% of the uncertain predictions eliminated. 4) CNN achieves the highest improvement among the three models with a 14\% improved accuracy ratio when 30\% of the uncertain predictions eliminated, while BERT only achieves 6\% improvement based on the accuracy with no uncertainty prediction eliminated. 5) Metric learning improves the accuracy by the largest ratio on CNN, but brings only a trivial performance improvement to BERT.

\begin{figure*}[b]
	\centering
	\subfigure[Metric Margin $m$ = 0.5]{%
		\label{fig:embed2}
		\includegraphics[trim=0.3cm 0.1cm 0.6cm 0.5cm,width=0.45\linewidth]{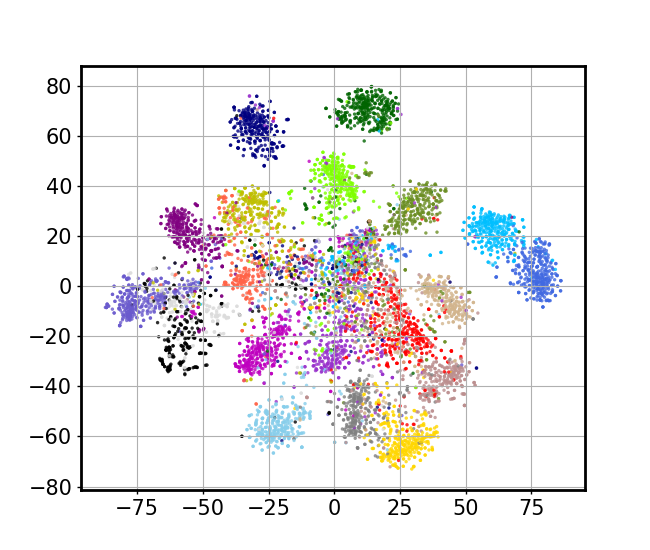}
	} %
	
	\subfigure[Metric Margin $m$ = 5]{%
		\label{fig:embed3}
		\includegraphics[trim=0.3cm 0.1cm 0.6cm 0.5cm,width=0.45\linewidth]{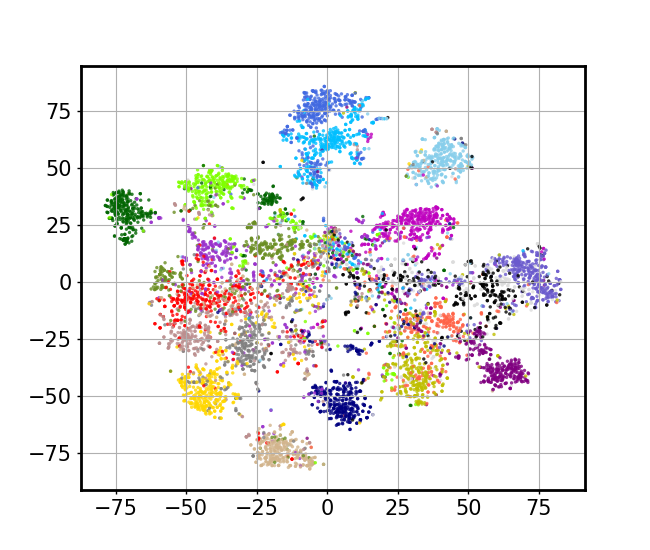}
	} %
	
	
	\caption{%
		\small Feature visualization of 20 NewsGroup testing data set in two dimensions by t-SNE algorithm.
	}%
	\label{fig:embedding_ap}
\end{figure*}

\section{Feature Visualization}

As the extended visualization to Figure \ref{fig:embedding}, Figure \ref{fig:embedding_ap} shows the 300-dimension feature representations for the 20 NewsGroup testing data set, with \ref{fig:embed2} - \ref{fig:embed3} presenting the features trained by metric learning with a margin parameterm=5 and 10. From the further experimental results, we can also get the conclusion that the distances between the inter-classes are clearly enlarged compared to the features trained without metrics.

\end{document}